\documentclass[11pt]{article}
\usepackage{color}
\usepackage{graphicx}
\usepackage{geometry}
\usepackage{latexsym}
\usepackage{amsfonts}
\usepackage{mathrsfs}
\usepackage{amssymb,amscd}
\usepackage[all]{xy}
\usepackage{amsmath}
\usepackage{fancyhdr}
\usepackage{fancybox}
\usepackage{xcolor}
\usepackage{array}
\usepackage{makecell}
\usepackage{multirow}
\usepackage{multicol}
\usepackage{slashbox}
\usepackage{indentfirst}
\usepackage{url}
 \author{Zhifeng Kong}
  \title{Convergence Analysis of the Dynamics of a Special Kind of Two-Layered Neural Networks with $\ell_1$ and $\ell_2$ Regularization}
\date{}

\begin{document}
\maketitle
\begin{abstract}
In this paper, we made an extension to the convergence analysis of the dynamics of two-layered bias-free networks with one $ReLU$ output. We took into consideration two popular regularization terms: the $\ell_1$ and $\ell_2$ norm of the parameter vector $w$, and added it to the square loss function with coefficient $\lambda/2$. We proved that when $\lambda$ is small, the weight vector $w$ converges to the optimal solution $\hat{w}$ (with respect to the new loss function) with probability $\geq (1-\varepsilon)(1-A_d)/2$ under random initiations in a sphere centered at the origin, where $\varepsilon$ is a small value and $A_d$ is a constant. Numerical experiments including phase diagrams and repeated simulations verified our theory.
\end{abstract}

\section{Introduction}
A substantial issue in deep learning is the theoretical analysis of complex systems. Unlike multi-layer perceptrons, deep neural networks have various structures \cite{summary}, which mainly come from intuitions, and they sometimes yield good results. On the other hand, the optimization problem usually turns out to be non-convex, thus it is difficult to analyze whether the system will converge to the optimal solution with simple methods such as stochastic gradient descent.

In Theorem 4 in \cite{related}, convergence for a system with square loss with $\ell_2$ regularization is analyzed. However, assumption 6 in \cite{related} requires the activation function $\sigma$ to be three times differentiable with $\sigma'(x)>0$ on its domain. Thus the analysis cannot be applied to some popular activation functions such as $ReLU$ \cite{relu} and $PReLU$ \cite{prelu}, where $ReLU(x)=\max(x,0)$ and $PReLU(x)=\max(x,\alpha x),\ 0<\alpha<1$.

Theorem 3.3 in \cite{work} provides another point of view to analyze the $\sigma=ReLU$ situation by using the Lyapunov method \cite{lyapunov}. The conclusion is weaker: the probability of convergence is less than $1/2$. However, this method successfully deals with this activation function. In this paper, we take into consideration $\ell_1$ and $\ell_2$ regularization and analyze the convergence of these two systems with an analogous method. Also, a similar conclusion is drawn in the end.

The square of the $\ell_1$ and $\ell_2$ norms of a vector $v$ are
\begin{equation}
\|v\|_1^2=\left(\sum_{i=1}^n |v_i|\right)^2,~~
\|v\|_2^2=\sum_{i=1}^n v_i^2=v^{\top}v.
\end{equation}
These two regularization terms are popular because they control the scale of $v$. Because there is an important difference between $\ell_1$ and $\ell_2$ regularization (usually it is possible to acquire an explicit solution of a system with $\ell_2$ regularization, but hard for a system with $\ell_1$ regularization), we need different tools to deal with the problems.

\section{Preliminary}
In this paper a two-layered neural network with one $ReLU$ output is considered. Let $X=(x_1,x_2,\cdots,x_N)^{\top}$, an $N\times d$ matrix ($N\gg d$), be the input data. Assume that the columns of $X$ are identically distributed Gaussian independent random $d$-dimensional vector variables: $x_i$'s $(i.i.d.)$ $\sim \mathcal{N}(0,I_d)$. Let $w$, a vector with length $d$, be the vector of weights (parameters) to be learned by the model. Let $w^*$ be the optimal weight with respect to $X$. Let $\sigma=ReLU$ be the activation function. Then, the output with input vector $x$ and weight $w$ is $g(x,w)=\sigma(x^{\top}w)$. For convenience, define $g(X,w)$ an $N\times1$ vector with $i^{th}$ element $g(x_i,w)$. Now, we are able to write down the loss function with the regularization term $R(w)$:
\begin{equation}E(w)=\frac{1}{2N}\|g(X,w^*)-g(X,w)\|^2+\frac{\lambda}{2}R(w),\end{equation}
where $\lambda\geq0$ is a parameter. When $\lambda=0$, there is no regularization. In this paper, we focus on the situation where $R(w)=$ $\|w\|_1^2$ or $\|w\|_2^2$ and $\lambda$ is very small.

We have a easy way to represent $g(X,w)$ by introducing a new matrix function $D$ given by $D(w)=diag\{d_1,d_2,\cdots, d_N\}$ where $d_i=1$ if $(Xw)_i>0$ and $d_i=0$ if $(Xw)_i\leq0$.  Then, $g(X,w)$ can be written in matrix form:
\begin{equation}g(X,w)=D(w)Xw.\end{equation}
Additionally, let $D^*=D(w^*)$ for convenience.

Now we introduce the gradient descent algorithm for the model. The iteration has the form
\begin{equation}w^{t+1}=w^t+\eta\Delta w^t,\end{equation}
where $\eta$ is the learning rate (usually small) and $\Delta w^t=-\nabla_w E(w^t)$ is the negative gradient of the loss function.
According to \cite{work} $\Delta w$ has the closed form
\begin{equation}\Delta w=\frac1N X^{\top}D(w)\left(D^*Xw^*-D(w)Xw\right)+\frac{\lambda}{2}\frac{\partial R}{\partial w}.\end{equation}
Its expectation (corresponding to $X$) is given explicitly by
\begin{equation}\mathbb{E}\Delta w=\frac 12(w^*-w)+\frac{1}{2\pi}\left((\alpha\sin\theta) w-\theta w^*\right)+\frac{\lambda}{2}\frac{\partial R}{\partial w},\end{equation}
where $\alpha=\|w^*\|/\|w\|$ and $\theta\in(0,\pi/2]$ is the angle between $w$ and $w^*$.

\section{Theoretical Analysis}

Usually, $w$ does not converge to $w^*$ because of the regularization term. Let $\hat{w}$ be the optimal weight vector that minimizes $E(w)$, i.e. $\frac{\partial E}{\partial w}(\hat{w})=0$. First, we'll solve $\hat{w}$ for small $\lambda$, and then we prove that $w^t$ will converge to $\hat{w}$ in $\hat{\mathcal{B}}_{\|w^*\|}(w^*)=\mathcal{B}_{\|w^*\|}(w^*)\setminus \ell(w^*)$ using the Lyapunov method \cite{lyapunov}, where $\mathcal{B}_{r}(y)=\{x\in\mathbb{R}^d:\|x-y\|^2\leq r^2\}$ and $\ell(y)$ is the line $\{ky:k\in\mathbb{R}\}$ .

We firstly provide three lemmas that help the analysis in sections 3.2 and 3.3. The lemmas show that extreme situations will happen with small probability, and provide some mathematical tricks that are useful in the theoretical analysis.

\subsection{Preparation}

\textbf{Lemma 1}: $A_k=Prob\left\{rank(D^*)\leq k\right\}=2^{-N}\sum_{i=0}^k\binom Ni$ for $k\in\{0,1,\cdots,N\}$.

\textit{Proof}: Let $x\sim\mathcal{N}(0,I_{d})$, then $Prob\{x^{\top}w^*\leq0\}=\frac12$. Thus,
\begin{equation}Prob\{rank(D^*)=i\}=\binom Ni \prod_{j=1}^i Prob\{x_j^{\top}w^*>0\}\cdot \prod_{j=i+1}^N Prob\{x_j^{\top}w^*\leq0\}=2^{-N}\binom Ni\end{equation}
Finally,
\begin{equation}A_k=\sum_{i=0}^k Prob\{rank(D^*)=i\}=2^{-N}\sum_{i=0}^k\binom Ni\end{equation}
When $N\gg k$, $A_k$ is a small value bounded by $2^{-N}k\binom Nk$. $\Box$

\textbf{Lemma 2}: $Prob\{X^{\top}D^*X \mbox{ is positive definite}\}=1-A_d$.

\textit{Proof}: First we show when $rank(D^*)>d$, $Prob\{X^{\top}D^*X\mbox{ is positive definite}\}=1$. Since $x_i$'s are $i.i.d.$, any $d$ rows of $X$ are linearly independent with probability 1. This implies that with probability 1 $Xr$ doesn't contain more that $d$ $0$'s $\forall r\in \mathbb{R}^d\setminus\{0\}$. However, $D^*$ has more than $d$ $1$'s, so $r^{\top}X^{\top}D^*Xr=(Xr)^{\top}D^*(Xr)>0\ a.s.$ Then since $Prob\{rank(D^*)>d\}=1-A_d$, the probability that $X^{\top}D^*X$ is positive definite also equals to this amount. $\Box$

\textbf{Lemma 3}: For a  positive definite matrix $B$ and a small value $\varepsilon$
\begin{equation}(B-\varepsilon I)^{-1}=(I+\varepsilon B^{-1}+\textbf{o}(\varepsilon))B^{-1}\end{equation}
where $\textbf{o}(\varepsilon)$ refers to a matrix with every element $=o(\varepsilon)$.

\textit{Proof}: Since $B$ is  positive definite, $B^{-1}$ exists. Then,
\begin{equation}\begin{array}{ll}
(B-\varepsilon I)^{-1}&=\left(B(I-\varepsilon B^{-1})\right)^{-1}\\
&=(I-\varepsilon B^{-1})^{-1}B^{-1}\\
&=(I+\varepsilon B^{-1}+\textbf{o}(\varepsilon))B^{-1}
\end{array}\end{equation}
This shows that $B^{-1}$ and $(B-\varepsilon I)^{-1}$ are closed to each other. $\Box$

\subsection{Convergence Area for the $\ell_2$ Regularization Case}
In this case, we have $R(w)=\|w\|_2^2=w^{\top}w$, and $\partial R/\partial w=2w$. Then, the loss function is given in the following equation:
\begin{equation}E(w)=\frac{1}{2N}\|g(X,w^*)-g(X,w)\|^2+\frac{\lambda}{2}\|w\|_2^2\end{equation}

\textbf{Theorem 1}: When $\lambda$ is small, $\hat{w}$ can be solved explicitly with probability $1-A_d$.

\textit{Proof}: Let $\partial E/\partial w=0$, and according to equation (5), we have
\begin{equation}X^{\top}D(\hat{w})(D^*Xw^*-D(\hat{w})X\hat{w})+\lambda N \hat{w}=0\end{equation}
Let's first assume that $D(\hat{w})=D^*$. Then the equation can be simplified as
\begin{equation}X^{\top}D^*Xw^*=(X^{\top}D^*X-\lambda NI_d)\hat{w}\end{equation}
Thus, we have
\begin{equation}\hat{w}=(X^{\top}D^*X-\lambda NI_d)^{-1}X^{\top}D^*Xw^*\end{equation}
The inverse exists with probability $1-A_d$ according to lemmas 2 and 3.

We now show that when $\lambda$ is small enough, this $\hat{w}$ ensures that $D(\hat{w})=D^*$. According to lemmas 2 and 3, we have
\begin{equation}\hat{w}=w^*+\lambda N(X^{\top}D^*X)^{-1}w^*+\textbf{o}_{d\times1}(\lambda)\end{equation}
It is sufficient to show that $X\hat{w}$ and $Xw^*$, two vectors in $\mathbb{R}^N$, share the same signs in the $N$ positions with probability 1. These two vectors are related by the equation
\begin{equation}X\hat{w}=Xw^*+\lambda NX(X^{\top}D^*X)^{-1}w^*+\textbf{o}_{N\times1}(\lambda)\end{equation}
Since $Xw^*$ doesn't contain 0 with probability 1, we can exclude these cases. Then, all terms after $Xw^*$ above don't influence the sign of $Xw^*$ when
\begin{equation}\lambda\leq\frac{1}{2N}\min_{1\leq i\leq N}\frac{|(Xw^*)_i|}{\left|(X(X^{\top}D^*X)^{-1}w^*)_i\right|}\end{equation}
The "2" on the denominator is used for eliminating the effects of $\textbf{o}_{N\times1}(\lambda)$. $\Box$

Now, we have shown that $\hat{w}$ is closed to $w^*$ when $\lambda$ is small. The next step is to show that $w$ converges to $\hat{w}$ in a certain area, which the Lyapunov method \cite{lyapunov} is very good at. In order to apply the Lyapunov method, we regard $t$ as a continuous index.

\textbf{Theorem 2}: With probability $1-A_d$, the following statement holds. When $N$ is large and $\lambda$ is small, consider the Lyapunov function $\mathcal{V}(w)=\frac12\|w-\hat{w}\|^2$.  We have $\dot{\mathcal{V}}(=\partial V/\partial t)<0$ in $\hat{\mathcal{B}}_{\|w^*\|}(w^*)$, and thus the system is asymptotically stable. That is, $w=w^{t}\rightarrow \hat{w}$ as $t\rightarrow\infty$.

\textit{Proof}: We can write $\dot{\mathcal{V}}$ as:
\begin{equation}\dot{\mathcal{V}}=(w-\hat{w})^{\top}\mathbb{E}\Delta w\end{equation}
In order to simplify, let $\hat{w}=w^*+\lambda T$, where $T$ is given by
\begin{equation}T=N(X^{\top}D^*X)^{-1}w^*+\textbf{o}_{d\times1}(1)\in\mathbb{R}^d.\end{equation}

Note $y=(\|w\|,\|w^*\|)^{\top}$; $\dot{\mathcal{V}}$ can be written as $-y^{\top}Ky$ where
\begin{equation}K=M+\lambda P\end{equation}
According to Lemma7.3 \cite{work}, $M$ is given by the following:
\begin{equation}M=\frac{1}{4\pi}\left(
      \begin{array}{cc}
        2\pi & -(2\pi-\theta)\cos\theta-\sin\theta \\
        -(2\pi-\theta)\cos\theta-\sin\theta & \sin2\theta+2\pi-2\theta \\
      \end{array}
    \right)
\end{equation}
$P$ can also be divided into two parts: $P=P_1+P_2$, where
\begin{equation}P_1=-\left(
      \begin{array}{cc}
        1 & -\frac{\cos\theta}{2} \\
        -\frac{\cos\theta}{2} & 0 \\
      \end{array}
    \right)\end{equation}
and $P_2$ satisfies that $y^{\top}P_2y=T^{\top}\mathbb{E}\Delta w$. From this, we see that $P$ is bounded. Since $M$ is  positive definite for $\theta\in(0, \pi/2]$ according to Lemma7.3 \cite{work}, when $\lambda$ is small, $K$ is also  positive definite for $\theta\in(0, \pi/2]$. As a result, $\dot{\mathcal{V}}<0$, which leads to the result that the system is asymptotically stable in $\hat{\mathcal{B}}_{w^*}(w^*)$. $\Box$

\subsection{Convergence Area for the $\ell_1$ Regularization Case}
In this case, we have $R(w)=\|w\|_1^2$, and $\partial R/\partial w=2\|w\|_1sign(w)$, where $sign(w)$ is the vector of signs of elements in $w$. Then, the loss function is given in the following equation:
\begin{equation}E(w)=\frac{1}{2N}\|g(X,w^*)-g(X,w)\|^2+\frac{\lambda}{2}\|w\|_1^2\end{equation}

\textbf{Theorem 3}: When $\lambda$ is small, $\hat{w}$ can be solved (not explicitly) with probability $1-A_d$.

\textit{Proof}: Let $\partial E/\partial w=0$, and according to equation (5), we have
\begin{equation}X^{\top}D(\hat{w})(D^*Xw^*-D(\hat{w})X\hat{w})+\lambda N \|\hat{w}\|_1sign(\hat{w})=0\end{equation}
We still assume that $D^*=D(\hat{w})$ to simplify the problem. Then, the equation becomes
\begin{equation}f(\lambda,w^*,\hat{w})=X^{\top}D^*Xw^*-X^{\top}D^*X\hat{w}+\lambda N \|\hat{w}\|_1sign(\hat{w})=0\end{equation}
This problem is hard to solve, so we use the Implicit Function Theorem \cite{implicit} here. The key is to examine whether the Jacobian matrix $J$ is invertible, where $J(i,j)=\partial f_i/\partial \hat{w}_j$. The result is
\begin{equation}\begin{array}{ll}
J(i,j)&=-(X^{\top}D^*X)_{ij}+\lambda N sign(\hat{w}_i)sign(\hat{w}_j)\\
&\displaystyle =-\sum_{k=1}^NI(x_k^{\top}w^*>0)x_{ki}x_{kj}+\lambda N sign(\hat{w}_i)sign(\hat{w}_j)
\end{array}\end{equation}
Since $X^{\top}D^*X$ is positive definite with probability $1-A_d$ according to Lemma 2, and when $\lambda$ is small the second term doesn't influence, we know that $J$ is then invertible. Thus, there exists a unique continuously differentiable function $g$ such that $\hat{w}=g(w^*,\lambda)$ is the solution. Notice that when $\lambda=0$, $\hat{w}=w^*$ is the solution. As a result, $\hat{w}=g(w^*,\lambda)$ can be extended as $w^*+\lambda u+\textbf{o}_{d\times1}(\lambda)$ for some vector $u$. Additionally, $u$ might be very large because there is an $N$ after $\lambda$ in equations (24)-(26).

Then, we show that for $\lambda$ small, we have $D(\hat{w})=D^*$. The analysis is quite similar to Theorem 1. When
\begin{equation}
\lambda\leq\frac12\min_{(Xw^*)_i>0}\frac{(Xw^*)_i}{|(Xu)_i|}
\end{equation}
we have that $D(\hat{w})=D^*$. $\Box$

\textbf{Remark}: In Theorem 3 the bound of $\lambda$ is given in equation (27), where there is an unknown vector $u$ on the denominator $(Xu)_i$. In fact, we are able to estimate its value from known quantities. When we apply the extension $\hat{w}=w^*+\lambda u+\textbf{o}(\lambda)$ to equation (25), we have
\begin{equation}
X^{\top}D^*X(-\lambda u+\textbf{o}(\lambda))+\lambda N\|\hat{w}\|_1sign(\hat{w})=0,
\end{equation}
which is equivalent to the following equation
\begin{equation}
X^{\top}D^*Xu=N\|\hat{w}\|_1sign(\hat{w})+\textbf{o}(1).
\end{equation}
As assumed in Theorem 3, $rank{D^*}=\delta>d$, and assume that $D^*_{ii}=1$ for $i=1,2,\cdots,\delta$. Let $X_{\delta}$ be the matrix consisting the first $\delta$ rows of $X$. Then, $X^{\top}D^*Xu=X_{\delta}^{\top}X_{\delta}u$. Thus, we have
\begin{equation}
u=N\|\hat{w}\|_1(X_{\delta}^{\top}X_{\delta})^{-1}sign(\hat{w})+\textbf{o}(1).
\end{equation}

Then we have
\begin{equation}
Xu = N\|\hat{w}\|_1X(X_{\delta}^{\top}X_{\delta})^{-1}sign(\hat{w})+\textbf{o}(1),
\end{equation}
which indicates that
\begin{equation}
\max_{(Xw^*)_i>0}(Xu)_i\leq 2N\|\hat{w}\|_1\|X(X_{\delta}^{\top}X_{\delta})^{-1}\|_{\infty}
\end{equation}
for small $\lambda$ such that $\|\hat{w}\|_1\leq(2-\epsilon)\|w^*\|_1$ with small value $\epsilon$ that eliminates the effect of $\textbf{o}(1)$ in equation (31). Finally, we are able to modify the bound in equation (27) by using the upper bound of $(Xu)_i$ in equation (32) to substitute this amount. The explicit bound is then given by the following equation:
\begin{equation}
\lambda\leq\frac{\min_{(Xw^*)_i>0}(Xw^*)_i}{4N\|w^*\|_1\|X(X_{\delta}^{\top}X_{\delta})^{-1}\|_{\infty}}.
\end{equation}
$\Box$

Although the explicit solution of $\hat{w}$ can't be found, we still draw the conclusion that $\hat{w}$ is closed to $w^*$ for small $\lambda$. This is enough for the Lyapunov method, because we are able to control $K$ in equation (20) with a similar way.

\textbf{Theorem 4}: The statement in Theorem 2 still holds for $\ell_1$ regularization.

\textit{Proof}: Similar to the analysis in Theorem 2, we still have equation (20) in this case with a different $P$. Thus, when $\lambda$ is small enough $K$ is positive definite, and the conclusion is still correct here. $\Box$

\subsection{The Final Result}
Since it's hard to draw samples from $\hat{\mathcal{B}}_{w^*}(w^*)$, we consider a small sphere $\mathcal{B}_0(r)$ centered at the origin. The analysis is in Theorem 7.4 (proof of Theorem 3.3) in \cite{work}.

\textbf{Theorem 5}: For both $\ell_1$ and $\ell_2$ regularization, if the initial weight vector $w^1$ is sampled uniformly in $\mathcal{B}_0(r)$ with $r\leq\varepsilon\sqrt{\frac{2\pi}{d+1}}\|w^*\|$, $w$ converges to $\hat{w}$ with probability $\geq\frac{1-\varepsilon}{2}(1-A_d)$.

\textit{Proof}: The proof is almost exactly the same as the proof of Theorem 7.4 in \cite{work}. The only thing to notice is that we exclude the line $\ell(w^*)$ because we need $\theta>0$. However, the line has measure zero and thus doesn't change the conclusion. $\Box$

Now, we have proved that Theorem 3.3 in \cite{work} still applies for $\ell_1$ and $\ell_2$ regularization with small $\lambda$. And this result is consistent with the argument that initial weights should be small rather than being large \cite{course}.

\section{Experiment Results and Analysis}
\begin{figure}[!h]
  \centering
  \includegraphics[width=3.5cm, height=2.5cm]{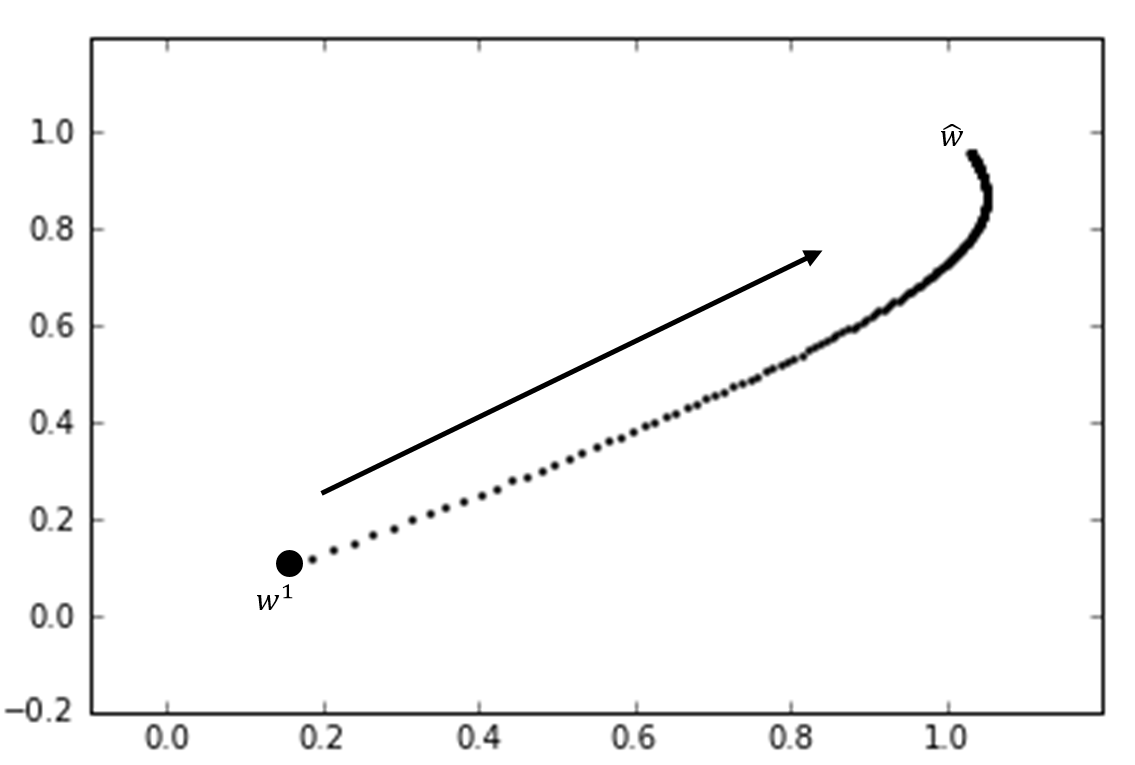}
  \includegraphics[width=3.5cm, height=2.5cm]{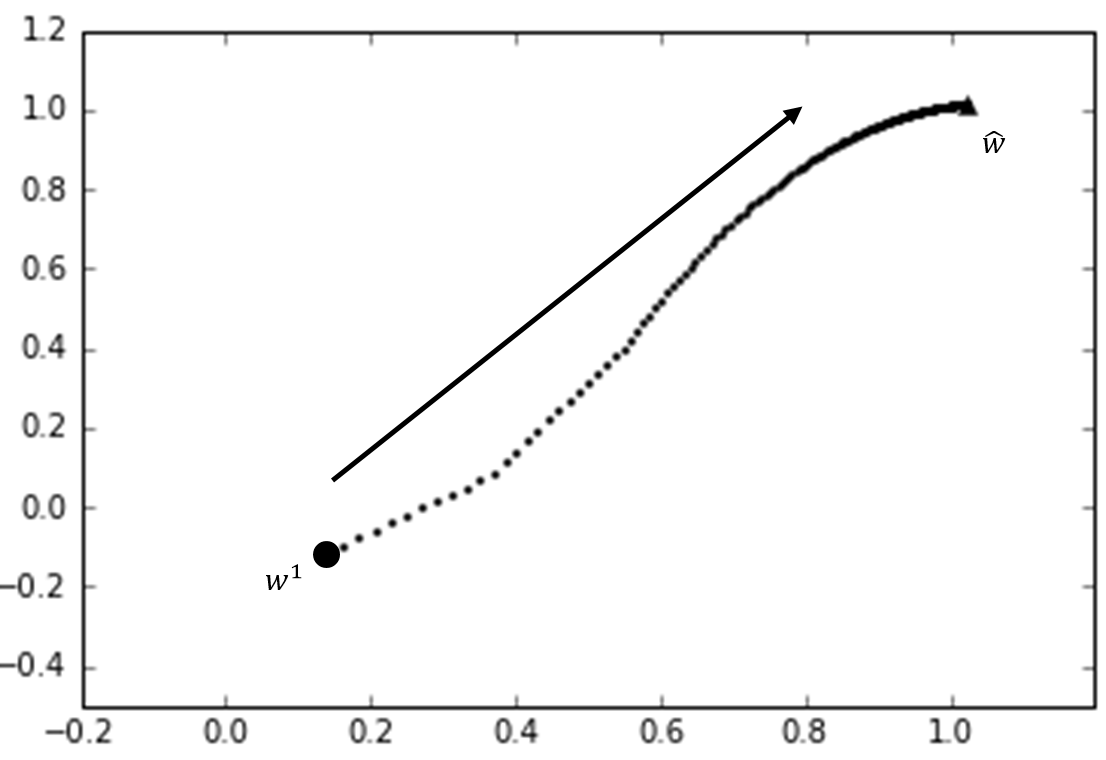}
  \includegraphics[width=3.5cm, height=2.5cm]{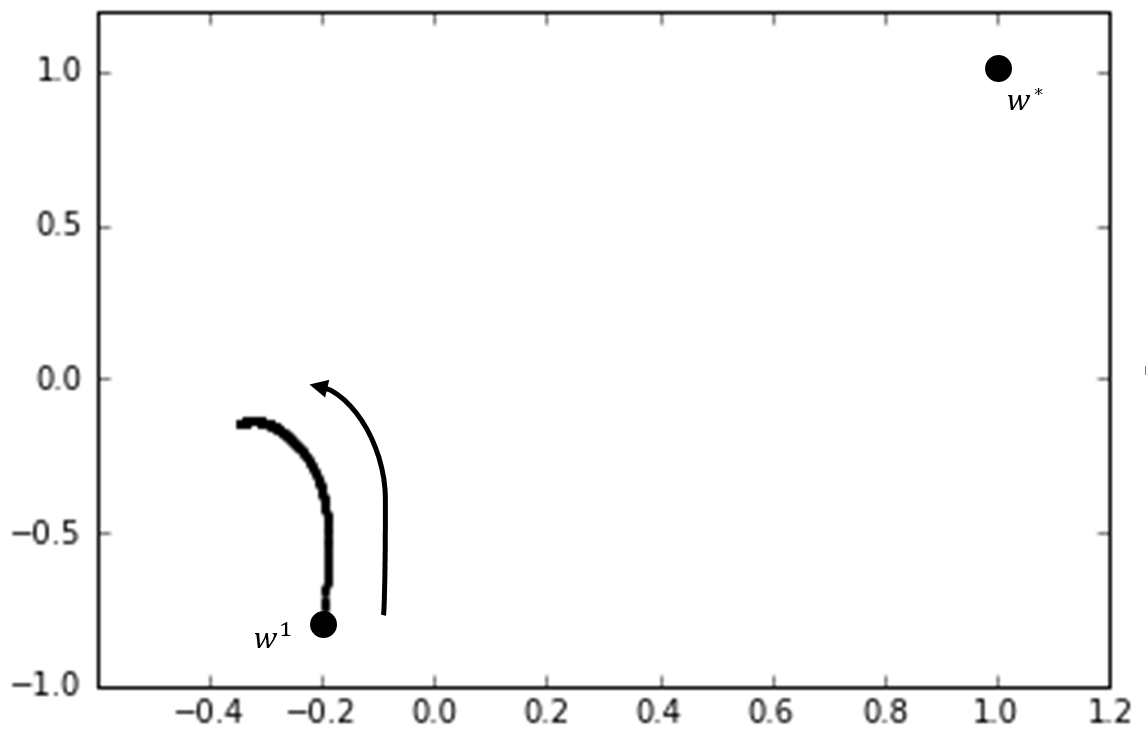}
  \includegraphics[width=3.5cm, height=2.5cm]{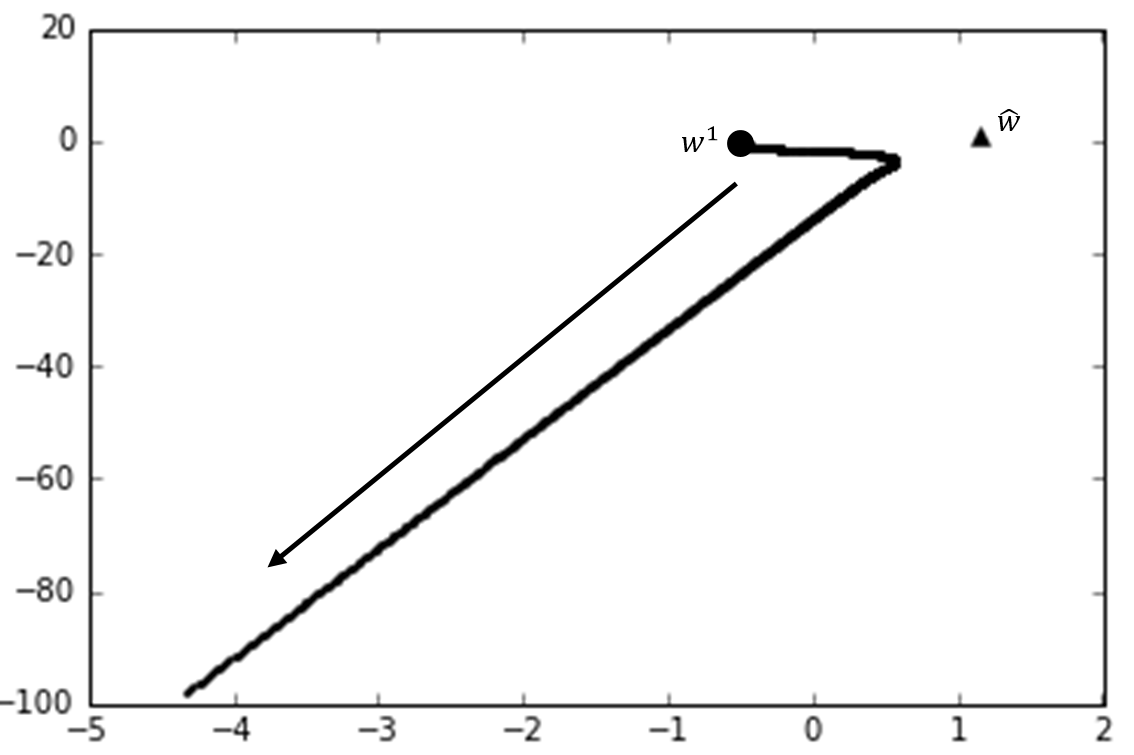}\\
  \caption{Four possible dynamics.
  The left 1 shows the dynamic that converges to $\hat{w}$ with $\ell_1$ regularization and $\lambda=0.01$.
  The left 2 shows the dynamic that converges to $\hat{w}$ with $\ell_2$ regularization and $\lambda=0.01$.
  The left 3 shows the dynamic that does not converge to $\hat{w}$ with $\ell_1$ regularization and $\lambda=0.1$.
  The left 4 shows the dynamic that does not converge to $\hat{w}$ with $\ell_2$ regularization and $\lambda=0.1$.
  }\label{four}
\end{figure}

\begin{figure}[!h]
  \centering
  \includegraphics[width=0.4\textwidth]{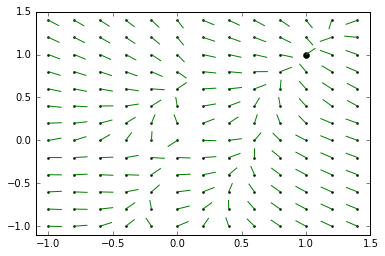}
  \includegraphics[width=0.4\textwidth]{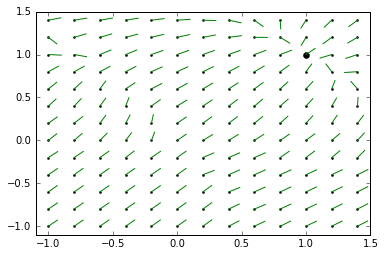}\\
  \caption{Phase diagrams (or vector fields, after normalized) in the $(x,y)$ plane of the dynamics with $\ell_1$(left) and $\ell_2$(right) regularization.}\label{phase}
\end{figure}

First, in Figure \ref{four} we demonstrate all possibilities: the dynamics converge/do not converge with $\ell_1$/$\ell_2$ regularization. The parameters are:
$N=10, d=2, \eta=0.05, \varepsilon=0.1,$ and $\lambda=0.01\mbox{(for cases with convergence)}, 0.1\mbox{(for cases that without convergence)}$.
In Figure \ref{phase} we show two phase diagrams (or vector fields, after normalized) of the dynamics with $\ell_1$ and $\ell_2$ regularization with randomly selected $X$ and parameters $N=10, d=2$, and $\lambda=0.01$. The big black point is $w^*=(1,1)$, the small black points are the grid points uniformly selected in the plane, and green lines refer to the orientations of $\delta w$ (from the end with a black point to the end without any point). Especially, when $w$ equals to $(0,0)$ in the $\ell_1$ case the dynamic is meaningless because $\partial R/\partial w$ does not exist.

Then, in order to examine the prediction given by Theorem 5, we made the following simulation. Under different values of $N$, $d$ and $\lambda$, we simulated the dynamics for 500 times and compared the experiment ratio of convergence to the theoretical ratio (that is, the probability) of convergence in Theorem 5. Specifically, for both $\ell_1$ and $\ell_2$ situation $N$ was selected in $\{10, 20, 100\}$, $d$ was selected in $\{2, 3, 5\}$, and $\lambda$ was selected in $\{0.001, 0.01, 0.1\}$. The learning rate $\eta$ was set to be 0.05 and $\varepsilon$ was set to be 0.1. Each time $X$ was sampled according to normal distribution and $w^1$ was sampled uniformly in $\mathcal{B}_0(r)$. The results for the $\ell_1$ and $\ell_2$ regularization case are demonstrated in Table \ref{result}.

\begin{table}[!h]
\begin{center}
\caption{The comparison between theoretical ratio of convergence (the 3rd col.) and experiment ratio of convergence (the 4th-9th col.) under different parameters.}\label{result}
\begin{tabular}{cc|c|ccc|ccc} \hline

\multirow{2}{*}{$d$}&\multirow{2}{*}{$N$}&\multirow{2}{*}{Theoretical}
&\multicolumn{3}{c|}{$\ell_2$ case with various $\lambda$}
&\multicolumn{3}{c}{$\ell_1$ case with various $\lambda$}\\
&&&0.001&0.01&0.1&0.001&0.01&0.1\\ \hline

\multirow{3}{*}2
&10&0.425&0.912&0.832&0.436&0.940&0.912&0.700\\
&20&0.450&0.992&0.976&0.578&0.970&0.956&0.840\\
&100&0.450&0.996&1&0.950&0.996&0.986&0.880\\ \hline
\multirow{3}{*}3
&10&0.373&0.852&0.712&{0.170}&0.966&0.972&0.736\\
&20&0.449&0.994&0.966&{0.342}&0.998&0.996&0.940\\
&100&0.450&1&1&0.856&1&1&0.962\\ \hline
\multirow{3}{*}5
&10&0.170&0.452&0.304&{0.016}&1&1&0.612\\
&20&0.441&0.97&0.820&{0.112}&1&1&0.960\\
&100&0.450&1&1&0.706&1&1&1\\ \hline

\end{tabular}
\end{center}
\end{table}

According to Table \ref{result}, we are able to make the following discussion.
$(i)$ As shown in the table, there are four bold numbers, all of which lie in the $\ell_2$ regularization case when $\lambda=0.1$, indicating that for the $\ell_2$ situation $0.1$ is beyond the upper bound of $\lambda$ for Theorem 1 or Theorem 2.
$(ii)$ In most situations, the experiment ratio of convergence decreases as $\lambda$ increases, and the gap between $\lambda=0.1$ and $0.01$ is much larger than the gap between $\lambda=0.01$ and $0.001$, which implies that $\lambda$ also plays an important role in the convergence probability in Theorem 5.
$(iii)$ In most cases the experiment ratio is much larger than the theoretical ratio. This indicates that outside the sphere $\hat{\mathcal{B}}_{\|w^*\|}(w^*)$ in Theorem 2 and Theorem 4 there is still much area in which the initial weights will converge to $w^*$.
$(iv)$ Under the same parameters, the experiment ratio of convergence in the $\ell_1$ case is always greater than that in the $\ell_2$ case. This shows that the $\ell_1$ regularization makes the dynamic easier to converge than the $\ell_2$ regularization does.

\section{Conclusion and Future Work}
In this paper, we presented our convergence analysis of the dynamics of two-layered bias-free networks with one $ReLU$ output, where the loss function includes the square error loss and $\ell_1$ or $\ell_2$ regularization on the weight vector. This is an extension to Theorem 3.3 in \cite{work}. We first solved the optimal weight vector $\hat{w}$ with small regularization coefficient for both cases, and then used the Lyapunov method \cite{lyapunov} to show that the system is asymptotically stable in certain area. In the final step, we claimed that Theorem 3.3 in \cite{work} is still correct in these two situations. We also verified our theory through numerical experiments including plotting the phase diagrams and making computer simulations.

Our work made a theoretical justification of convergence for two popular models. We started from the intuition that small regularization doesn't change the system too much, and our conclusion is compatible with this intuition. In the future, we plan to analyze the system with larger regularization, since in real situations $\lambda$ is fixed to be, for example, 0.5, which may be larger than the bound in equations (17) and (27). This is more difficult since we won't expect $D^*=D(\hat{w})$, and other advanced techniques may be applied. We also plan to consider other popular regularization terms, and provide a more general theory on this topic.

\end{document}